\documentclass[letterpaper, 10 pt, conference]{ieeeconf}  

\IEEEoverridecommandlockouts                              
\usepackage{soul}
\usepackage{graphicx}
\usepackage{flushend}

\title{\LARGE \bf
FACT: A Full-body Ad-hoc Collaboration Testbed \\for Modeling Complex Teamwork
}

\author{Gopika Ajaykumar$^{1,*}$, Annie Mao$^{2}$, Jeremy Brown$^{2}$, and Chien-Ming Huang$^{1}$\\%
Johns Hopkins University
\thanks{This work was supported in part by the National Science Foundation Graduate Research Fellowship Program under Grant No. DGE-1746891 and the Nursing/Engineering joint fellowship from the Johns Hopkins University.}\thanks{$^1$Department of Computer Science, Johns Hopkins University}
\thanks{$^2$Department of Mechanical Engineering, Johns Hopkins University}
\thanks{
        {\tt\small $^*$gopika@cs.jhu.edu}}}

\begin{document}

\maketitle
\thispagestyle{empty}
\pagestyle{empty}

\begin{abstract}
Robots are envisioned to work alongside humans in applications ranging from in-home assistance to collaborative manufacturing. Research on human-robot collaboration (HRC) has helped develop various aspects of social intelligence necessary for robots to participate in effective, fluid collaborations with humans. However, HRC research has focused on dyadic, structured, and minimal collaborations between humans and robots that may not fully represent the large scale and emergent nature of more complex, unstructured collaborative activities. Thus, there remains a need for shared testbeds, datasets, and evaluation metrics that researchers can use to better model natural, ad-hoc human collaborative behaviors and develop robot capabilities intended for large scale emergent collaborations. We present one such shared resource---FACT (Full-body Ad-hoc Collaboration Testbed), an openly accessible testbed for researchers to obtain an expansive view of the individual and team-based behaviors involved in complex, co-located teamwork. We detail observations from a preliminary exploration with teams of various sizes and discuss potential research questions that may be investigated using the testbed. Our goal is for FACT to be an initial resource that supports a more holistic investigation of human-robot collaboration. 
\end{abstract}

\section{Introduction}
Collaboration is a fundamental process that enables humans to perform various complex activities. For example, in professional automobile racing, pit crews swiftly replace tires, refuel, and carry out necessary repairs and adjustments within highly limited time spans; surgeons, nurses, anesthetists, and assistants work together seamlessly to perform surgeries; and firefighters respond collectively to a wide range of emergencies and catastrophes. Beyond professional endeavors, day-to-day interactions also frequently involve people working together, such as when people coordinate actions and distribute efforts while assembling furniture. 

In many of these instances, collaboration emerges on an ad-hoc basis, with teammates communicating and determining task roles and actions spontaneously. This emergent collaboration allows humans to effectively initiate and shape team-based behaviors according to their real-time task needs. Achieving similar emergent collaboration in human-robot teaming and interaction is critical for applications such as in-home assistance, flexible manufacturing, and search-and-rescue, where collaborative actions require high adaptability. However, much of the prior work on human-robot collaboration has focused on prescribed scenarios where collaborators have predefined, static roles and on dyadic human-robot collaborations in tabletop settings (e.g., \cite{huang2016anticipatory, chen2018planning, baraglia2017efficient, dragan2015effects, hayes2016autonomously}). Furthermore, the task scenarios used in HRC research often involve minimal collaborative activity, such as hand-offs from robots to humans, providing a limited view of how collaboration may need to evolve in larger scale interactions.

Moving towards developing shared testbeds, datasets, and evaluation metrics focused on large scale, emergent human-robot collaboration can help drive HRC research towards investigating the more unstructured, ad-hoc, and involved collaborative activities that have so far remained largely unexplored. In this work, we present \emph{Full-body Ad-hoc Collaboration Testbed} (FACT), a testbed that researchers can use to understand human behaviors in emergent collaboration and develop robot capabilities based on these behaviors. We provide implementation details to enable researchers to recreate FACT, describe observations from a preliminary exploration using the testbed, and discuss future steps to further drive research into developing flexible, emergent human-robot collaborations.

\begin{figure}[t]
  \includegraphics[width=3.3in]{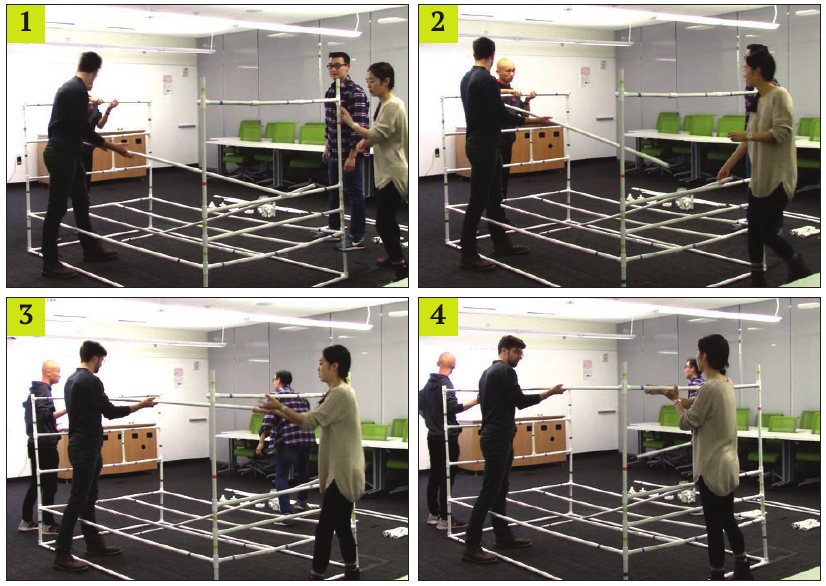}
  \caption{FACT (Full-body Ad-hoc Collaboration Testbed) enables investigation of emergent collaborative behaviors in complex assembly, such as dynamic sub-team formation.}
  \label{fig:teaser}
\end{figure}

\begin{figure*}[t]
  \includegraphics[width=\textwidth]{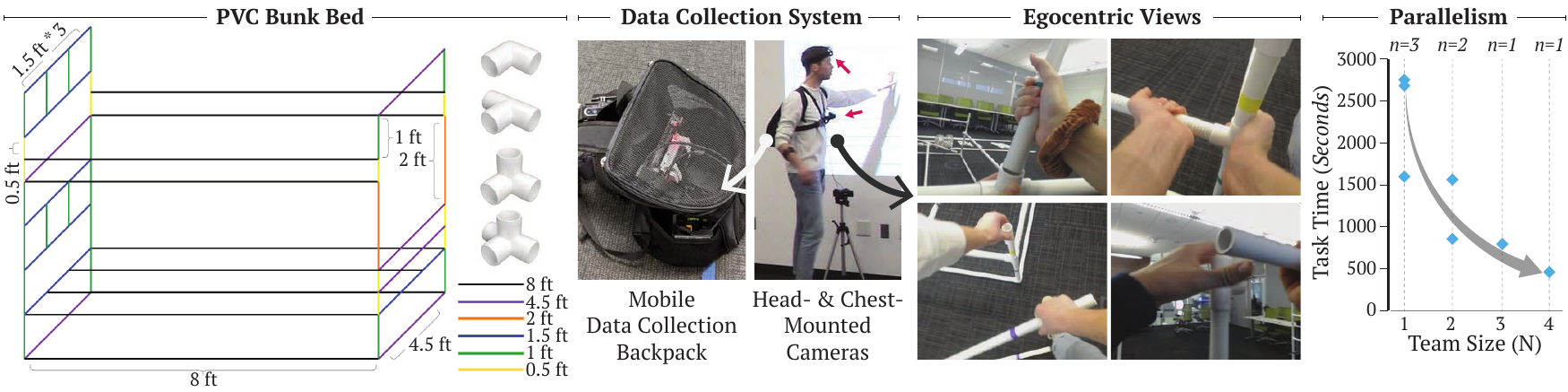}
  \caption{FACT includes a PVC bunk bed consisting of variously sized pipes and connectors of different forms (\emph{left}) and a mobile data collection backpack with first-person sensors that can capture dynamic, egocentric views during collaborative assembly (\emph{middle}). Our preliminary exploration using the testbed involved teams of various sizes, which allowed us to observe how team size can enhance task efficiency through mechanisms such as collaborative parallelism (\emph{right}).}
  \label{fig:bunkbed-data}
\end{figure*}

\section{FACT: Full-body Ad-hoc Collaboration Testbed}
We developed FACT as a testbed focused on a collaborative assembly scenario where teammates construct a ``bunk bed" using PVC pipes (Fig. \ref{fig:teaser}). The bunk bed structure is 8 ft. by 4.5 ft., is 6 ft. tall, and consists of pipes and connectors of different sizes and forms (Fig. \ref{fig:bunkbed-data}, left). The bunk bed's structure opens multiple possibilities in terms of sequencing steps and assigning collaborative roles during assembly, which enables the study of unstructured, emergent collaborative behaviors. Due to the bed's size, its assembly can involve bigger teams of people compared to the pairwise collaborations conventionally examined in HRC research. It involves natural opportunities for participants to work together (e.g., moving a long pipe) and offers opportunities to engage in collaborative parallelism, wherein team members work on sub-tasks that can be executed in parallel to complete a joint task. Furthermore, its size and form allow researchers to capture rich interaction dynamics in full-body collaboration. We note that the bunk bed affords natural cooperative activities, contrasting with experimental task scenarios used in prior relevant research, which either did not necessarily require collaboration (i.e., the task could be easily completed by one person) (e.g., \cite{knepper2014distributed}) or used small scale tabletop tasks focused on cognitive collaboration in lieu of full-body collaboration (e.g., \cite{shah2010empirical}). 

To facilitate exploration of the cues and user behaviors involved in our collaborative assembly scenario, we include a mobile data collection backpack in FACT (Fig. \ref{fig:bunkbed-data}, middle)\footnote{The software and parts list for FACT are available at https://github.com/intuitivecomputing/FACT.}. The backpack consists of a Nvidia Jetson TX2, a battery, and sensors, including a head-mounted camera, a chest-mounted camera, and a Myo EMG sensor. The backpack can be easily extended to work with additional or different sensors as needed. We run ROS on the Jetson TX2 to collect and synchronize sensor data. This backpack enables capture of egocentric views, which prior work has suggested can enable user-friendly human-robot collaborative assembly that better reflects user preferences \cite{wang2020see}, and dynamic information on collaboration as participants move around and work together during a task. In addition, we include three stationary video recorders to capture the process of collaboration. 

\begin{figure*}[t]
\centering
\includegraphics[width=\textwidth]{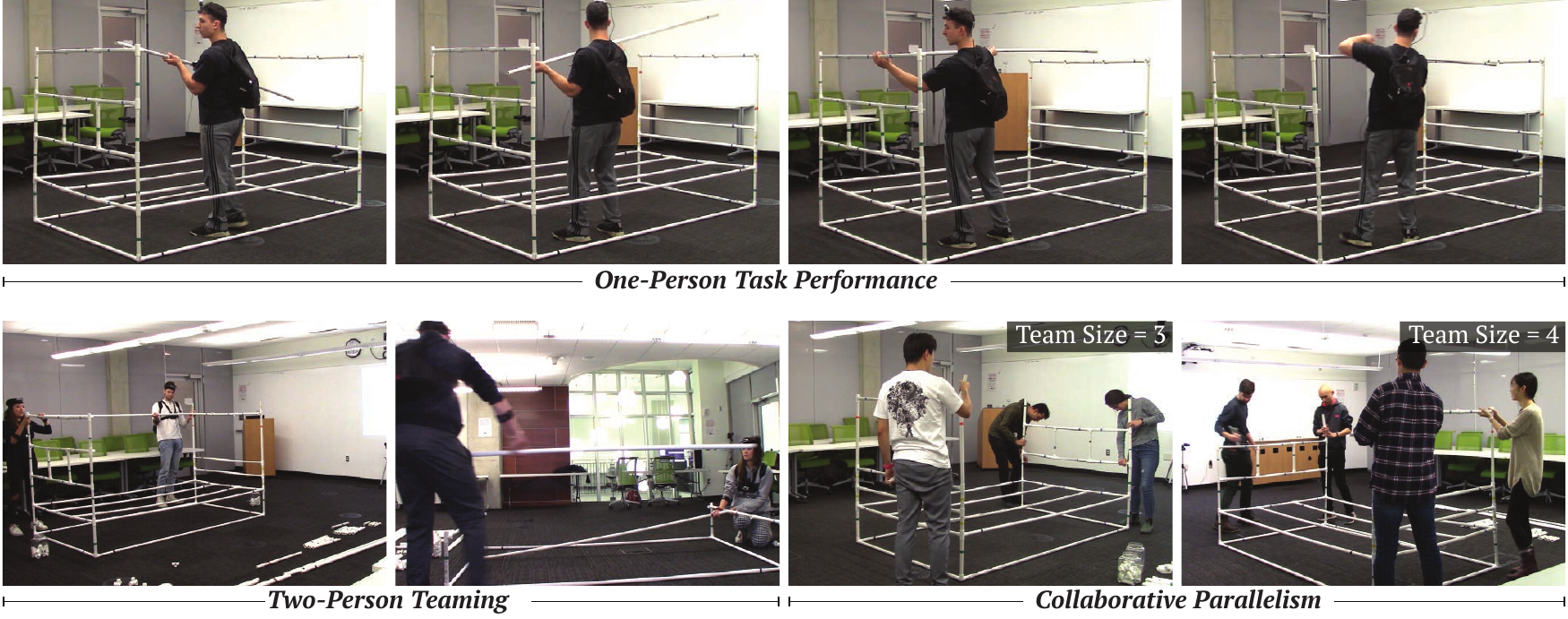}
\caption{We observed different behaviors and strategies corresponding to different team sizes. One-person task performance often involved difficulties, especially when handling bigger parts (\emph{top}), whereas multi-person team performance minimized such difficulties through teaming and collaborative parallelism (\emph{bottom}).}
\label{fig:various-teaming}
\end{figure*}

\section{Preliminary Exploration of Full-Body Ad-Hoc Collaboration}
To date, we have conducted a preliminary exploration of large scale emergent collaboration using FACT. Study participants were provided with an illustration of the assembled bunk bed (Fig. \ref{fig:bunkbed-data}, left), which was projected on a wall so that they could easily access and refer to it. Furthermore, the bunk bed parts were freely available in a central location in the task environment such that participants could determine how they wanted to manage and allocate the parts among team members. As a result, participants were free to choose how to assemble the structure, allowing us to study emergent collaboration (e.g., how participants agreed on assembly order and on who did what). Our preliminary exploration thus far has involved three one-person teams, two two-person teams, one three-person team, and one four-person team completing the experimental task (Fig. \ref{fig:various-teaming}). Below, we highlight key observations from our exploration.

\begin{itemize} \itemsep0em
    \item \textbf{Collaborative parallelism}. One common theme we observed was that participants worked in parallel on different aspects of the task. Parallelism was observed in two-person, three-person, and four-person collaborations. Moreover, a larger team (more than two people) tended to break into two-person sub-teams, which seemed to be a basic unit of collaboration (Fig. \ref{fig:various-teaming}). Sub-teams were dynamic and fluid, as the teaming was reconfigured from time to time.
    \item \textbf{Direct and indirect requests for help}. A common trigger for sub-team formations was explicit or implicit requests for help. Participants used explicit verbal requests, such as ``Can you pass me that connecting joint?", or implicitly signaled the need for help through inadequate task performance (Fig. \ref{fig:teaser}, 1 \& 2).
    \item \textbf{Multimodal communication}. Participants naturally employed multimodal behaviors (e.g., gaze cues, gestural illustration, and verbal requests) when communicating and coordinating with each other. This observation was in line with the body of literature on multimodal communication in human interaction \cite{clark1996using,mcneill1992hand}.     Multimodal behaviors served as communicative vehicles that allowed team members to synchronize their mental models of a task, facilitating effective teamwork.
    \item \textbf{Benefits of teamwork}. We observed various difficulties that participants faced in the one-person task performance compared to the collaborative task performance (Fig. \ref{fig:various-teaming}, top). Additionally, our preliminary data indicated that teamwork improved task efficiency (Fig. \ref{fig:bunkbed-data}, right). Team members were able to complement each other's actions and regularly participated in collaborative parallelism to speed up the assembly progress.
    \item \textbf{Flexible task plan}. There were many viable plans of execution (e.g., different assembly sequences) to assemble the bunk bed structure. We observed that participants did not develop and commit to a rigid plan of execution in detail before carrying out the task; instead, they maintained a flexible, approximate task plan and collaboratively determined immediate next steps by communicating as the task unfolded. This observation mirrored previous findings indicating that flexibility is key to efficient human teamwork, which benefits from quick adaptation and error recovery \cite{suchman1987plans}.     \item \textbf{Distributed planning}. We also observed that teams relied on distributed, rather than centralized, planning in emergent teamwork. In particular, there was no single leader commanding the team, and task planning happened throughout the course of assembly without team members committing to a specific, rigid task plan. Furthermore, the teams' plans were not necessarily the optimal task plan in terms of assembly sequence and task parallelism. In other words, while participants' task plans may be locally optimal in terms of immediate collaborative outcomes, they are not always globally optimal in the context of overall task performance metrics such as efficiency.    \item \textbf{Error handling}. Errors are common in complex collaborations. We observed two kinds of errors: 1) incorrect task assembly (e.g., using the wrong pieces) and 2) communication breakdowns (e.g., misunderstanding which piece was requested by a partner). Participants mitigated (e.g., apologizing or laughing after an error) and repaired (e.g., disassembling and reassembling or resynchronizing mental models through consultation with teammates) errors to ensure the success of the task.
    \item \textbf{Team dynamics}. We observed---especially in two-person teams---that some participants would take on a leadership role, giving more direct commands, while others acted as more passive partners during the collaboration. Additionally, we found that participants engaged in both functional (task-relevant) and social (e.g., making jokes) communication. While it may not contribute directly to task completion, social communication is crucial to team harmony \cite{rousseau2006teamwork}. 
\end{itemize}

These observations lay the foundation for investigating complex human-robot emergent collaboration. For example, they pave the way for further investigation into the following research questions. \begin{itemize}\itemsep0em     \item \textbf{Synchronization of mental models}. \emph{When and how do team members synchronize their mental models of the joint task as the emergent collaboration unfolds? Are there differences in the way people perform intra- and inter- sub-team synchronization and coordination? How may members collectively commit to an execution plan through multimodal communication?} Answers to these questions will help inform the development of computational representations and methods needed to enable grounded large scale human-robot collaboration.
    \item \textbf{Collaboration through parallelism and task assistance}. \emph{How should robot partners productively participate in emergent collaboration? When should a robot assist its human partners, and when should it work on parallel tasks? When and how do people form and leave sub-teams? Is sub-team formation based on teammates' proximity to the task area or other factors such as task needs and teaming history?} An effective robot will need to act at appropriate times to dynamically pair with different partners when needed and depart from sub-teams to make progress on its individual sub-task.
        \item \textbf{Error identification, mitigation, and recovery}. \emph{How can a robot recognize and prevent its task errors? How do team members identify communication breakdowns?} \emph{How can a robot identify when it is unable to recover from errors and requires human assistance? How do team dynamics influence error recovery?} Although prior work has explored how multimodal human responses to robot actions may be indicative of robot errors and their severity \cite{stiber2020not}, further research is needed to investigate when and how a robot should ask for help, as well as how to foster positive team dynamics to minimize task errors and communication breakdowns.
\end{itemize}

\section{Discussion}
To achieve the full potential of human-robot teaming in a wider range of activities, HRC research must move beyond small scale, static collaborations to focus on large scale, dynamic collaborations. Shared testbeds, datasets, and evaluation metrics derived from large scale emergent collaboration scenarios can help researchers methodically implement and evaluate the robot behaviors needed to achieve complex, ad-hoc human-robot collaborations. In this work, we presented FACT (Full-body Ad-hoc Collaboration Testbed), which consists of a PVC bunk bed collaborative assembly scenario and an accompanying mobile data collection setup for researchers to better understand and model individual and team behaviors during emergent collaboration and to develop and evaluate collaborative robot capabilities. 

Unlike previous testbeds and datasets for modeling human-human interaction (e.g., \cite{liu2016data, vogt2017system,  glasauer2010interacting, kontogiorgos2018multimodal}), our testbed enables the capture of natural interactions that do not require participants to role-play or act in specific collaborative roles, allows modeling of bigger team behaviors beyond dyadic interactions, and involves full-body collaborations that move beyond tabletop settings. Furthermore, while human-robot interaction researchers have explored various aspects of social intelligence for HRC, such as multimodal understanding (e.g., \cite{whitney2016interpreting, matuszek2014learning}), action coordination (e.g., \cite{breazeal2004teaching}), task parallelism (e.g., \cite{roncone2017transparent}), moderation of team dynamics (e.g., \cite{jung2015using, traeger2020vulnerable}), flexible task planning (e.g., \cite{hayes2016autonomously}), and action anticipation (e.g., \cite{koppula2015anticipating}), our testbed allows for a more holistic investigation into how an interplay of these various aspects contributes to large scale emergent collaborations and may also provide insight into designing interaction conventions for complex emergent collaboration (e.g., \cite{han2020structuring}).

We contribute an openly accessible testbed for investigating large scale emergent collaboration in this work. Our future work will involve the development of a shared dataset from collaborations using FACT, which will include full-body and egocentric manipulation data, similar to \cite{kratzer2020mogaze} but focused on team-based collaborative activity rather than on one-person task performance. We would also like to add to the existing set of established evaluation metrics for human-robot collaboration (e.g., \cite{hoffman2019evaluating}) to capture aspects of human-robot interaction specific to large scale emergent collaboration, such as dynamic sub-team formation. To minimize the influence of limited manipulation and motion capabilities of real-word robots on research on emergent collaboration, we would also like to develop a simulation counterpart to FACT that enables researchers to deploy and test behaviors on virtual AI agents in the bunk bed assembly scenario. Overall, we hope that FACT can serve as an initial tool that encourages the development of additional resources and research directions that can advance investigation of social intelligence for large scale emergent collaboration.

\bibliographystyle{IEEEtran.bst}
\bibliography{2021-icra-ajaykumar-preprint}

\end{document}